\documentclass{article}
\usepackage{ijcai21}

\usepackage{times}

\usepackage{soul}
\usepackage{url}
\usepackage[hidelinks]{hyperref}
\usepackage[utf8]{inputenc}
\usepackage[small]{caption}
\usepackage{graphicx}
\usepackage{amsmath}
\usepackage{booktabs}

\usepackage{multirow}
\usepackage{amssymb}

\title{Temporal Spatial-Adaptive Interpolation with Deformable Refinement for Electron Microscopic Images}

\author{
Zejin Wang$^{1,2}$\footnote{Contact Author}\And
Guodong Sun$^{1,2}$\And
Lina Zhang$^2$\and
Guoqing Li$^2$\and
Hua Han$^{1,3,4}$\\
\affiliations
$^1$Institute of Automation, Chinese Academy of Sciences\\
$^2$School of Artificial Intelligence, University of Chinese Academy of Sciences\\
$^3$School of Future Technology, University of Chinese Academy of Sciences\\
$^4$Center for Excellence in Brain Science and Intelligence Technology\\
\emails
\{wangzejin2018, sunguodong2019, lina.zhang, guoqing.li, hua.hua\}@ia.ac.cn
}

\begin{document}

\maketitle

\begin{abstract}
Recently, flow-based methods have achieved promising success in video frame interpolation. However, electron microscopic (EM) images suffer from unstable image quality, low PSNR, and disorderly deformation. Existing flow-based interpolation methods cannot precisely compute optical flow for EM images since only predicting each position's unique offset. To overcome these problems, we propose a novel interpolation framework for EM images that progressively synthesizes interpolated features in a coarse-to-fine manner. First, we extract missing intermediate features by the proposed temporal spatial-adaptive (TSA) interpolation module. The TSA interpolation module aggregates temporal contexts and then adaptively samples the spatial-related features with the proposed residual spatial adaptive block. Second, we introduce a stacked deformable refinement block (SDRB) further enhance the reconstruction quality, which is aware of the matching positions and relevant features from input frames with the feedback mechanism. Experimental results demonstrate the superior performance of our approach compared to previous works, both quantitatively and qualitatively.
\end{abstract}

\section{Introduction}
Electron microscopic (EM) image interpolation aims at employing the temporal information between consecutive frames to increase the z-axis resolution and produce better continuity. For example, through interpolating 4nm z-axis resolution images, we can get 2nm imaging effects. Since high-resolution serial slices in the z-axis contain fine motion dynamics, this approach contributes to analyzing the continuous structure in biological tissues.

In recent years, deep convolutional neural networks have been exploited for video frame interpolation and shown promising effectiveness. Prior works~\cite{niklaus2017adaptive,niklaus2017separable} estimate the spatial adaptive convolution kernel for each pixel and further use separable strategies to reduce model capacity. Simply extracting adaptive kernels also accounts for their poor performance under complex scenarios. Benefiting from the deep optical flow estimation~\cite{dosovitskiy2015flownet,ilg2017flownet}, subsequent works~\cite{niklaus2018context,bao2019memc,DAIN} precisely model the inter-frame motion relationship in video frame interpolation and produce visually pleasing results. However, deep optical flow estimation only predicts a unique offset for each position and warps the pixel at that position with the corresponding offset. Limited by its modeling capabilities, flow-based interpolation methods perform poorly on complex EM images, with severe smoothing and artifacts.

To overcome the above problems, we propose an effective temporal spatial-adaptive interpolation network (TSAIN) for EM images based on feature pyramid structure and deformable convolution. Specifically, the feature temporal and spatial-adaptive (TSA) interpolation module first extracts the pyramid temporal features under deformable convolution guidance and then adaptively samples the spatial-related features through the proposed residual spatial adaptive block (RSAB). Taking interpolated features as a reference, inaccuracy predicted intermediate features inevitably magnify the error once we warp the input features. Therefore, we further propose a deformable refinement block (DRB) performing feedback adjustment under the supervision of input features. We also notice that the feedback deformable correction is stackable. Increasing the number of stacked modules can refine more accurate intermediate features.

\begin{figure*}[ht]
\centering
\includegraphics[width=\textwidth]{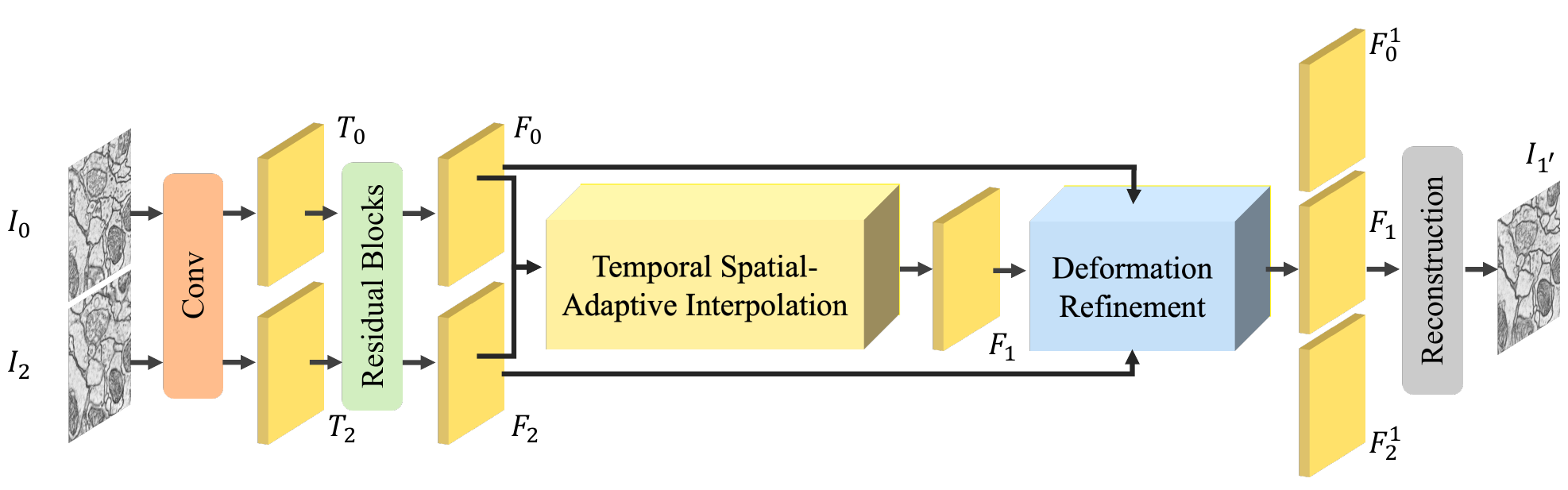}
   \caption{Overview of our temporal spatial-adaptive interpolation framework. This framework utilizes feature temporal spatial-adaptive interpolation to aggregate temporal and spatial-related features, then refines more precise interpolated features under the supervision of input features in a feedback mechanism.}
\label{fig:overview}
\end{figure*}

The contributions in this paper can be summarized as follows:

\begin{enumerate}
\item We propose a simple but effective EM image interpolation framework. The proposed model generates intermediate features with temporal spatial-adaptive sampling and further refine more accurate interpolated features through a feedback operation.
\item We propose a stacked deformable refinement module to obtain the best relevant features under the supervision of input frames. 
\item Experimental results show that our approach achieves state-of-the-art results on the EM benchmark dataset, superior to the recent frame interpolation algorithms.
\end{enumerate}

\section{Related Work}
\subsection{Video Frame Interpolation}
Video frame interpolation (VFI) aims to predict non-existent intermediate frames through input frames.~\cite{long2016learning} first introduce general convolutional neural networks (CNN) into video frame interpolation. As directly synthesizing interpolated frames by the CNN, severe artifacts and blurriness invariably occur.~\cite{liu2017video} proposes the deep voxel flow to warp the input frames based on triple sampling, which suffers low blurriness but performs poorly in sceneries with substantial motion.~\cite{niklaus2017adaptive,niklaus2017separable} proposes to replace voxel flow with adaptive convolutions, which synthesize pixels from a large neighborhood. Benefiting from the deep optical flow~\cite{ilg2017flownet}, explicitly estimating dynamic motion using optical flow estimation becomes an indispensable step in video frame interpolation~\cite{bao2019memc,DAIN}.~\cite{bao2019memc} integrates adaptive convolution and motion estimation into a single model achieving promising results with motion compensation. Part of the work~\cite{DAIN} explicitly detects the occlusion by exploring the depth information, which gains robust results even meets occlusion. However, optical flow estimation only predicts an offset for each coordinate. This single-coordinate single-offset mechanism severely restricts the optical flow modeling ability in more complex scenarios. For example, in a complex electron microscope scene, the image quality is unstable, the signal-to-noise ratio is low, and the deformation is anisotropic and disordered.

\subsection{Slice Interpolation}
Early researchers propose several slice interpolation methods to extract precise deformation fields, including morphological methods~\cite{lee2000morphology} and registration methods~\cite{penney2004registration}. These conventional approaches are mainly based on the crucial assumption that structural variations among slices are sufficiently small, making the above methods not suitable for sparsely sampled slices. Recently, there are some CNN-based slice interpolation methods.~\cite{afshar2018carisi} employs a simple encoding and decoding structure to perform binary image interpolation. ~\cite{wu2020inter} enhances the accuracy of medical image segmentation by predicting intermediate frames. However, the approach only adopts adaptive convolution to predict the interpolated frame, which is not suitable for images with visible motion or even complex EM images.

\subsection{Video Super-Resolution}
Video super-resolution (VSR) intends to reconstruct high-resolution (HR) video frames from the corresponding low-resolution (LR) frames and adjacent LR frames. One primal problem lies in how to perform the temporal alignment with adjacent LR frames. Some VSR approaches~\cite{sajjadi2018frame,xue2019video} use optical flow for explicit alignment, first estimate the displacement field between adjacent frames and the reference LR frame, and then take the predicted motion for warping the supporting frames. However, it is difficult to estimate the flow map accurately, and warping operations also introduce artifacts and additional errors. Especially on EM images, which are more complex than video frames, flow-based methods can hardly estimate accurate motion. Some VSR methods~\cite{wang2019edvr} replace optical flow estimation with deformable convolution~\cite{dai2017deformable,zhu2019deformable} for implicit temporal alignment to solve this problem. These methods estimate multiple offsets for each position and design a modulate mechanism. Therefore, the VSR method based on deformable convolution has a stronger modeling ability and anti-interference ability than optical flow methods.

\section{Proposed Method}
Given two input EM frames $\mathbf{I}_{0}$ and $\mathbf{I}_{2}$, which are continuous in the z-axis, our goal is to synthesize the corresponding intermediate frame $\hat{\mathbf{I}}_{1}$. To accurately extract the deformation field from the complex EM images and deal with the unstable image quality, we propose a novel temporal and spatial-adaptive interpolation framework, which aggregates temporal content and spatial-related information progressively. We first encode the input feature maps: $\mathbf{F}_{0}$ and $\mathbf{F}_{2}$, using the feature extractor with a convolutional layer and $k_1$ residual blocks~\cite{he2016deep}. Furthermore, to better leverage temporal information and feedback mechanisms, we introduce a stackable deformable refinement module to process consecutive feature maps: $\{\mathbf{F}_{k}\}^{2}_{k=0}$. Increasing the number of stacked modules, the proposed refinement modules generate more precise interpolation features. Finally, we reconstruct the intermediate frames from the refined features with $k_{2}$ stacked residual blocks. The overall structure is illustrated in Figure~\ref{fig:overview}. 

\begin{figure}[ht]
\centering
\includegraphics[width=0.45\textwidth]{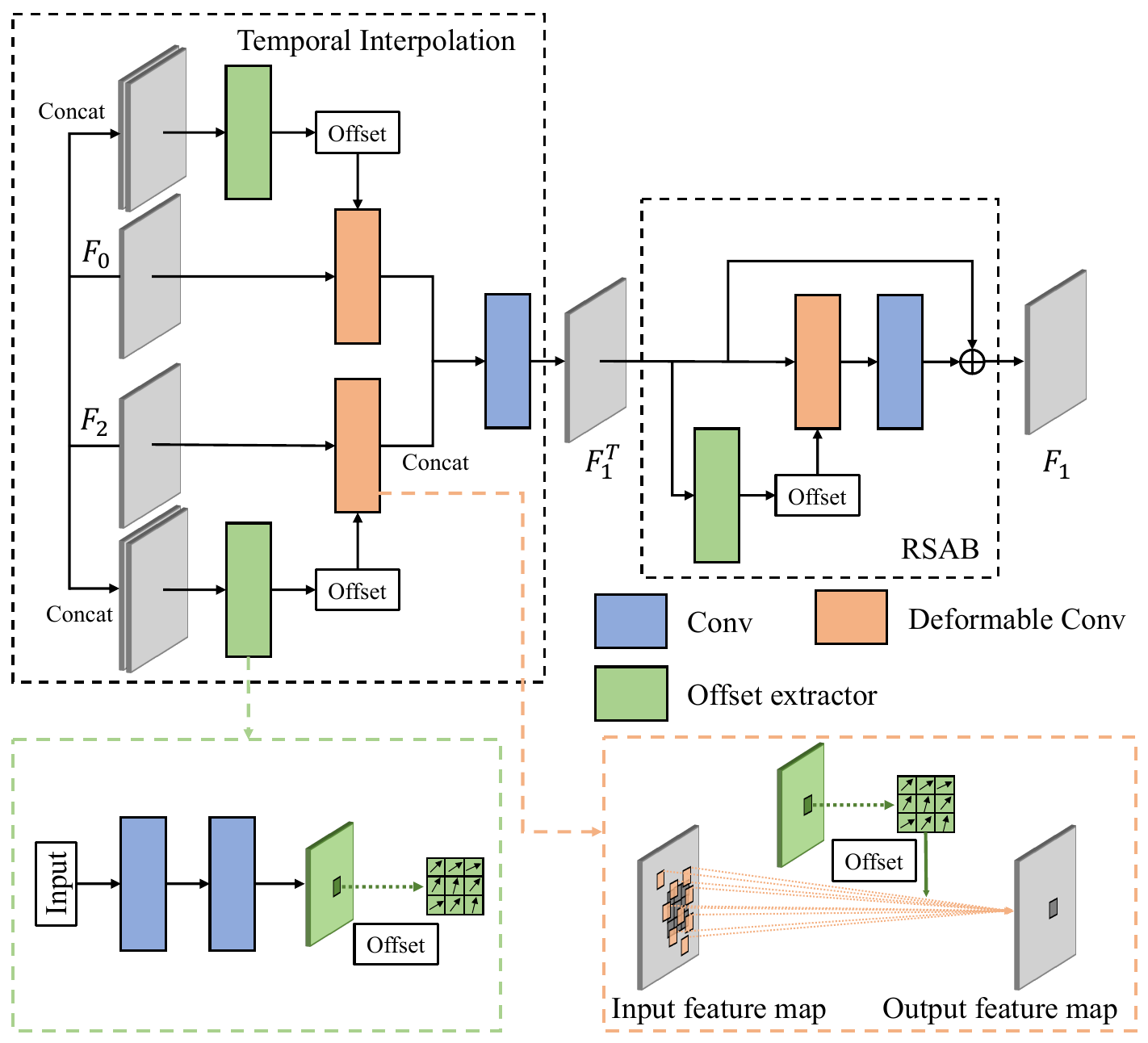}
   \caption{Frame feature temporal spatial-adaptive interpolation based on deformable convolution.}
\label{fig:tsa}
\end{figure}

\subsection{Temporal Spatial-Adaptive Interpolation}
Given input feature maps extracted from the input frames, our goal is to generate the missing feature map corresponding to the intermediate frame. Recent interpolation methods use deep optical flow to estimate motion, which leads to poor performance on complex EM images with large motion and unstable image quality. Unlike previous methods, we divide the synthesis of interpolation features into two steps (see Figure~\ref{fig:tsa}): (1) the temporal interpolation module $T(\cdot)$ synthesizes the temporal interpolated features; (2) The proposed spatial-adaptive module $S(\cdot)$ further sample spatial-related information based on the extracted temporal features. The general form of interpolation function $f(\cdot)$ can be formulated as:
\begin{equation}
F_{1} = f(F_{0}, F_{2}) = S(B(T_{0}(F_{0}, \Theta_{0}),T_{2}(F_{2},\Theta_{2})),\Phi),
\end{equation}
where $T_{0}(\cdot)$ and $T_{2}(\cdot)$ are two temporal sampling functions and $S(\cdot)$ is a saptial-adaptive sampling function; $\Theta_{0}, \Theta_{2}$ and $\Phi$ are the corresponding sampling parameters; $B(\cdot)$ is a fusion function to fuse two temporal interpolated features.

\subsubsection{Temporal Interpolation}
Inspired by the recent flow-based method~\cite{DAIN}, which first estimates the bi-directional optical flow, i.e., $f_{0\rightarrow2}$ and $f_{2\leftarrow0}$, and then approximate the intermediate flows, i.e., $f_{1\rightarrow0}$ and $f_{1\rightarrow2}$, we propose to achieve a similar process with deformable convolution~\cite{dai2017deformable,zhu2019deformable} implicitly. 

Specifically, $T_{0}(\cdot)$ implicitly combines the processes of capturing the forward motion $f^{m}_{0\rightarrow2}$ and approximating the intermediate motion $f^{m}_{1\rightarrow0}$ into a single step, denoted by the corresponding sampling parameter $\Theta_{0}$. Similarly, $T_{2}(\cdot)$ implicitly combines the processes of capturing the backward motion $f^{m}_{2\rightarrow0}$ and approximating the intermediate motion $f^{m}_{1\rightarrow2}$ into one step, denoted by the corresponding sampling parameter $\Theta_{2}$. 

Here, we take the temporal sampling function $T_{0}(\cdot)$ as an example. It uses $F_{0}$ and $F_{2}$ as input to predict the corresponding sampling parameter $\Theta_{0}$ for sampling $F_{0}$:

\begin{equation}
\Theta_{0}=p_{0}([F_{0},F_{2}]),
\end{equation}
where $\Theta_{0}$ refers to a learnable offset generated by several convolution layers $p_{0}$; $[,]$ denotes the channel-wise concatenation. With $\Theta_{0}$ and $F_{0}$, the temporal interpolated feature $F_{0\rightarrow1}$ can be computed by the modulated deformable convolution~\cite{zhu2019deformable}:

\begin{equation}
F_{0\rightarrow1} = T_{0}(F_{0}, \Theta_{0})=DConv(F_{0}, \Theta_{0}),
\end{equation}

Similarly, we learn an offset $\Theta_{2}=p_{2}([F_{2},F_{0}])$ as the corresponding sampling parameter, and then generate the temporal interpolated feature $F_{2\rightarrow1}=T_{2}(F_{2}, \Theta_{2})$.

To aggregate the final temporal interpolated feature $F^{T}_{2}$, we use a simple fusion function $B(\cdot)$:

\begin{equation}
F^{T}_{2}=B(F_{0\rightarrow1}, F_{2\rightarrow1})=Conv([F_{0\rightarrow1}, F_{2\rightarrow1}]),
\end{equation}
where $Conv(\cdot)$ is a $1\times1$ convolution operation. Since we use the final temporal interpolated feature $F^{T}_{2}$ to predict the missing intermediate frame, it will enforce the synthesized temporal feature to be close to the real intermediate feature. Therefore, the two temporal sampling parameters $\Theta_{0}$ and $\Theta_{2}$ will implicitly learn to estimate the forward and backward motion information, which is similar to bidirectional optical flow.

\subsubsection{Residual Spatial-Adaptive Blocks}
Influenced by the unstable EM image quality, such as unstable contrast, brightness, and blur, we notice that the intermediate frames generated using only temporal context show partial missing in membrane structure and discontinuity in edges. To alleviate this problem, we propose a spatial-adaptive sampling function $S(\cdot)$ to adapt to spatial
texture changes on the $F^{T}_{1}$, which further strengthens the accuracy of generated interpolation features. 

For simplicity,  $S(\cdot)$ denotes a residual spatial-adaptive block (RSAB). The spatial-adaptive module takes the $F^{T}_{1}$ as input to predict the sampling parameter:

\begin{equation}
\Phi = g(F^{T}_{1}),
\end{equation}
where $\Phi$ is a learnable offset; $g$ denotes several convolution layers. With the learned offset, the RSAB can be formulated as:
\begin{equation}
F_{1}=f_{RSAB}(F^{T}_{1})=S(F^{T}_{1},\Phi),
\end{equation}
here, $F_{1}$ denotes the final intermediate frame feature map generated by our temporal spatial-adaptive module. More specifically, we have:
\begin{equation}
f_{RSAB}(F^{T}_{1})=Conv(Act(DConv(F^{T}_{1}, \Phi)))+F^{T}_{1},
\end{equation}
where $DConv$ denotes the modulated deformable convolution~\cite{zhu2019deformable}; $Act$ is the activation function (leaky ReLU); $Conv$ is the traditional convolution. The architecture of RSAB is shown in Figure~\ref{fig:tsa}. To strengthen the performance of $S(\cdot)$, we cascade three residual spatial adaptive blocks.

With exploring temporal context and spatial-related information, our temporal and spatial-adaptive interpolation can handle large chaotic motion and unstable image quality in EM images.

\begin{figure}[ht]
\centering
\includegraphics[width=0.3\textwidth]{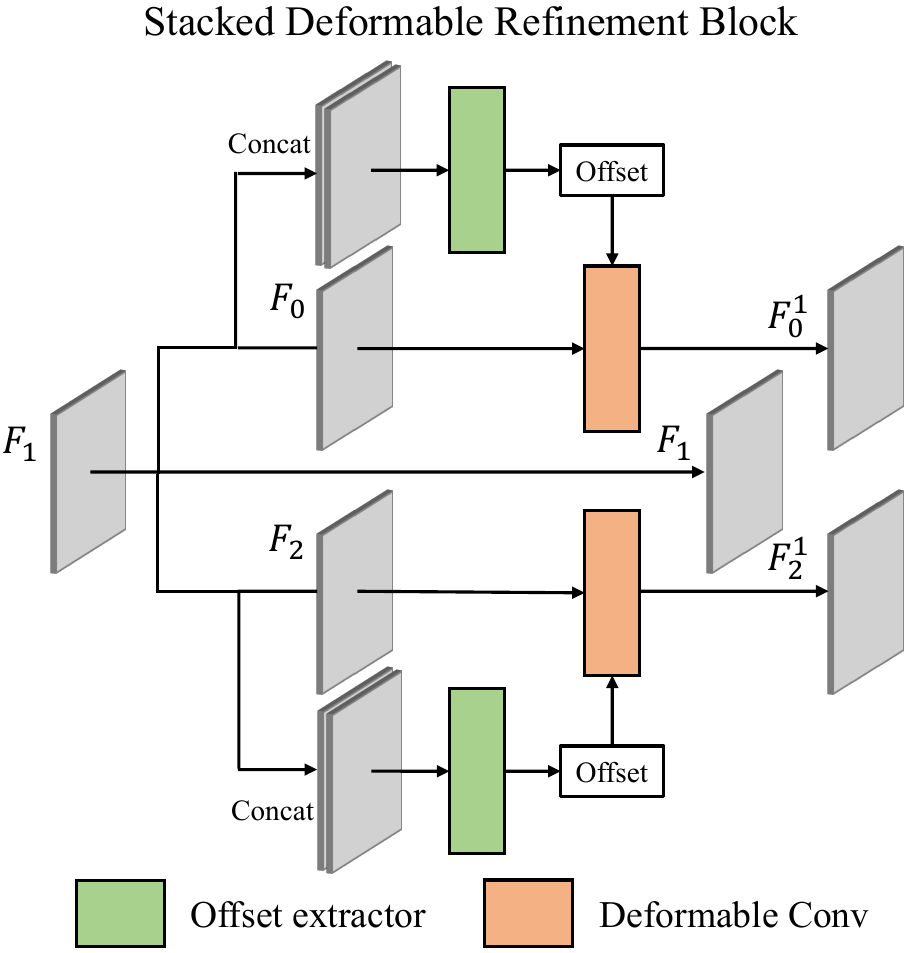}
   \caption{Stacked Deformable Refinement Block based on deformable convolution.}
\label{fig:DRBs}
\end{figure}

\subsection{Stacked Deformable Refinement Blocks}
Now we have the missing intermediate feature map $F_{1}$ for generating the corresponding intermediate frame $\hat{I}_{1}$.
Meanwhile, we have also got the input frame feature maps $F_{0}$ and $F_{2}$. Recent reference-based super-resolution method~\cite{shim2020robust} has proved that extracting aligned relevant features from a reference image contributes to increasing the performance of single image super-resolution. Therefore, we aggregate the aligned relevant features from the input frames with deformable convolution~\cite{zhu2019deformable}, rather than reconstructing an intermediate frame from the corresponding interpolated feature map. Taking the input frame features as a reference, if the extracted intermediate frame features are accurate enough, we further refine the alignment features beneficial to reconstructing the intermediate frame. Conversely, the inaccurate intermediate frame features inevitably amplify the reconstruction error since the intermediate frame features are obtained from the input frames.

Because of the feedback mechanism, the proposed deformable refinement block (DRB) further improves the performance of the intermediate frame reconstruction under input frames' supervision. The DRB (see Figure~\ref{fig:DRBs}) is defined as follows:
\begin{equation}
F^{1}_{0}, F_{1},F^{1}_{2}=f_{DRB}(F_{0},F_{1},F_{2}),
\end{equation}
where $F^{1}_{0}$ is the aligned relevant feature extracted from $F_{0}$; $F^{1}_{2}$ is the aligned relevant feature extracted from $F_{2}$. Taking $F^{1}_{0}$ as an example, we define how to extract alignment related features from $F_{0}$ as:
\begin{equation}
F^{1}_{0}=f_{dcn}(F_{0},\Phi_{0}),
\end{equation}
where $\Phi_{0}=\{\Delta p_{k}, \Delta m_{k}|k=1,\dots,K\}$ is a learnable offset for input frame feature map $F_{0}$; $f_{dcn}(\cdot)$ denotes the modulated deformable convolution~\cite{zhu2019deformable}. The $\Phi_{0}$ can be formulated as:
\begin{equation}
\Phi_{0}=g_{0}([F_{0},F_{1}]),
\end{equation}
Here, $g_{0}$ denotes a general function of several convolution layers; $[,]$ is a channel-wise concatenation operator.

More specifically, for each position $p$ on the aligned relevant feature map $F^{1}_{0}$, we have:
\begin{equation}
F^{1}_{0}(p)=\sum^{K}_{k=1} w_{k}\cdot F_{0}(p + p_{k} + \Delta p_{k})\cdot \Delta m_{k},
\end{equation}
where $k$ and $K$ correspondingly denote the index and the number of kernel weights. $w_{k}$, $p_{k}$, $\Delta p_{k}$ and $\Delta m_{k}$ are the $k$-th kernel weight, the $k$-th fixed offset, the learnable offsets for the $k$-th location and the learnable modulation scalar, respectively.  

Similarly, we learn an offset $\Phi_{2}=g_{2}(F_{2},F_{1})$ corresponding to the input frame feature map $F_{2}$, and then generate the aligned relevant feature map $F^{1}_{2}=f_{dcn}(F_{2}, \Phi_{2})$.

We notice that increasing the number of deformable refinement modules improves the performance of generated intermediate frames. To extract accurate aligned relevant feature maps, we use three stacked deformable refinement blocks as follows:

\begin{equation}
F^{1}_{0}, F_{1},F^{1}_{2}=f_{DRB}(f_{DRB}(f_{DRB}(F_{0},F_{1},F_{2}))),
\end{equation}

In the experiment, the three DRBs share the same network structure but have different weights. The next DRB takes the output of the previous DRB as input.

\subsection{Loss Functions}
The overall loss function to train the model is defined as:
\begin{equation}
\label{eq:loss}
L_{total}=\lambda_{pixel}L_{pixel} + \lambda_{perc}L_{perc}+\lambda_{style}L_{style},
\end{equation}
where $\lambda_{pixel}, \lambda_{perc}$ and $\lambda_{style}$ are the weights for pixel-wise loss, perceptual loss and style loss, respectively. In the experiment, we set $\lambda_{pixel}=\lambda_{perc}=1$ and $\lambda_{style}=10^{6}$.

\paragraph{Pixel-wise loss.}
The pixel-wise loss aims at reducing the divergence between systhesized intermediate frames and ground truth. The pixel-wise loss is definned as:
\begin{equation}
L_{pixel}=\frac{1}{HW}\sum^{H-1}_{i=0}\sum^{W-1}_{j=0}||\hat{I}_{1}(i,j)-I^{gt}_{1}(i,j)||^{2}_{2},
\end{equation} 

\begin{table*}[ht]
\centering
\resizebox{\textwidth}{!}{
\begin{tabular}[b]{*{12}{c}}
\toprule
\multirow{2}*{Methods} &
    \multicolumn{3}{c}{cremi$\_$triplet A} &
    \multicolumn{3}{c}{cremi$\_$triplet B} &
    \multicolumn{3}{c}{cremi$\_$triplet C} &
    Params & Runtime \\
    & PSNR & SSIM & IE & PSNR & SSIM & IE & PSNR & SSIM & IE & (Million) & (s) \\
\midrule
\midrule
DVF~\cite{liu2017video} & 18.83 & 0.4457 & 21.95 & 17.11 & 0.3360 & 26.63 & 16.89 & 0.3456 & 27.27 & 3.82 & 0.390 \\
SepConv~\cite{niklaus2017separable} &  17.52 & 0.4095 & 25.89 & 16.32 & 0.3522 & 28.26 & 16.07 & 0.3454 & 29.82 & 21.7 & \textbf{0.159}  \\
ToFlow~\cite{xue2019video} & 19.05 & 0.4920 & 21.41 & 17.28 & 0.3916 & 25.91 & 17.31 & 0.4046 & 25.80 & \textbf{1.44} & 0.402 \\
DAIN~\cite{DAIN} & 16.78 & 0.4264 & 28.20 & 15.67 & 0.3460 & 31.85 & 15.24 & 0.3210 & 33.59 & 24.0 & 0.553 \\
\midrule
TSAIN (ours) & \textbf{19.42} & \textbf{0.5140} & \textbf{20.46} & \textbf{17.52} & \textbf{0.4042} & \textbf{25.29} & \textbf{17.90} & \textbf{0.4347} & \textbf{24.15} & 5.71 & 0.232 \\
\bottomrule
\end{tabular}}
\caption{Quantitative comparisons with the state-of-the-art methods on cremi$\_$triplet datasets. The proposed TSAIN algorithm significantly outperforms other methods in terms of PSNR, SSIM and IE.}
\label{tab:cmps}
\end{table*}

\paragraph{Perceptual loss.}
Perceptual loss~\cite{johnson2016perceptual} aims at mitigating the blurriness casused by the pixel-wise loss, which contributes to generate more realistic results. The perceptual loss is defined in the feature level:
\begin{equation}
\label{eq:perc}
L_{perc}=\sum^{R-1}_{r=0}\frac{\Psi^{\hat{I}_{1}}_{r}-\Psi^{I_{1}}_{r}}{N_{\Psi^{I_{1}}_{r}}},
\end{equation}
where $\Psi$ is the $16$-layer VGG~\cite{simonyan2014very} network pretrained on ImageNet, and we use layer $relu1\_2$, $relu2\_2$, $relu3\_3$ and $relu4\_3$. Specifically, $\Psi^{I_{1}}_{r}$ is the activation from the $r$th layer of VGG given the input $I_{1}$, and $N_{\Psi^{I_{1}}_{r}}$ is the number of elements in the $r$th layer.

\paragraph{Style loss.}
We also introduce the style loss~\cite{johnson2016perceptual}, which aims to keep the image style for style transfer. Style loss is similar to perceptual loss, and can be defined as:
\begin{equation}
L_{style}=\sum^{R-1}_{r=0}\frac{1}{C_{r}C_{r}}\frac{|(\Psi^{\hat{I}_{1}}_{r})^{T}(\Psi^{\hat{I}_{1}}_{r})-(\Psi^{I_{1}}_{r})^{T}(\Psi^{I_{1}}_{r})|}{C_{r}H_{r}W_{r}},
\end{equation}
where $H_{r}, W_{r}, C_{r}$ is the shape of the VGG feature as in perceptual loss.

\subsection{Implementation Details}
{}In our implementation, $k_1=5$ and $k_2 = 40$ residual blocks are used in feature extraction and frame reconstruction modules, respectively. We randomly crop a triplet of EM image patches with the size of $256 \times 256$ and take out the odd-indexed $2$ frames as inputs, and the corresponding frame as supervision. For data augmentation, we randomly rotate $90^{\circ}$, $180^{\circ}$ and $270^{\circ}$, horizontally flip and randomly inverse their{} temporal order. We adopt a Pyramid, Cascading and Deformable (PCD) structure in~\cite{wang2019edvr} to employ temporal deformable alignment and apply Adam~\cite{kingma2014adam} as our optimizer. The initial learning rate is $4 \times 10^{-4}$ and decay by a factor of 0.1. The batch size is set to be $8$ and trained on one Tesla V$100$ GPU for $100$ epochs.

\section{Experiment}

\subsection{Datasets and Metrics}
\paragraph{Datasets.}
We evaluate the proposed approach on the EM images. However, there is no ready-made dataset for the EM image interpolation task that is currently being developed. Here, we use the CREMI dataset provide by MICCAI 2016 Challenge at~\href{https://cremi.org/}{https://cremi.org/}, which is divided into three sub-datasets named padder A, B, and C, respectively. Taking CREMI’s padder version A as an example, we first convert A dataset with the hdf5 format into png format to get 200 images with a resolution of $3072\times3072$. After that, we adopt the template matching algorithm to match three consecutive images selected from it. We then traverse from left to right from top to bottom in 512 steps and crop the three consecutive images with a resolution of $512\times512$ after alignment, and save them as a sample. Finally, we delete all samples with defects, weak continuity, and substantial differences in blurring. The processed CREMI datasets are named as cremi$\_$triplet A, cremi$\_$triplet B and cremi$\_$triplet C, respectively. Each dataset adopts a triplet as a sample for training, where each triplet contains three consecutive EM images with a resolution of $512 \times 512$. To eliminate the brightness inconsistency of the EM images, we perform histogram specifications on each dataset, which improves the robustness of the interpolation algorithms. 

\paragraph{Metrics.}
The average Interpolation Error (IE), Peak Signal to Noise Ratio (PSNR), Structural Similarity Index (SSIM), model parameters (Params) and run time (Runtime) are adopted to evaluate the performance of different methods. In particular, PSNR and SSIM are calculated on grayscale EM images. Moreover, lower IE indicates better performance.

\begin{figure}[ht]
\centering
\includegraphics[width=0.48\textwidth]{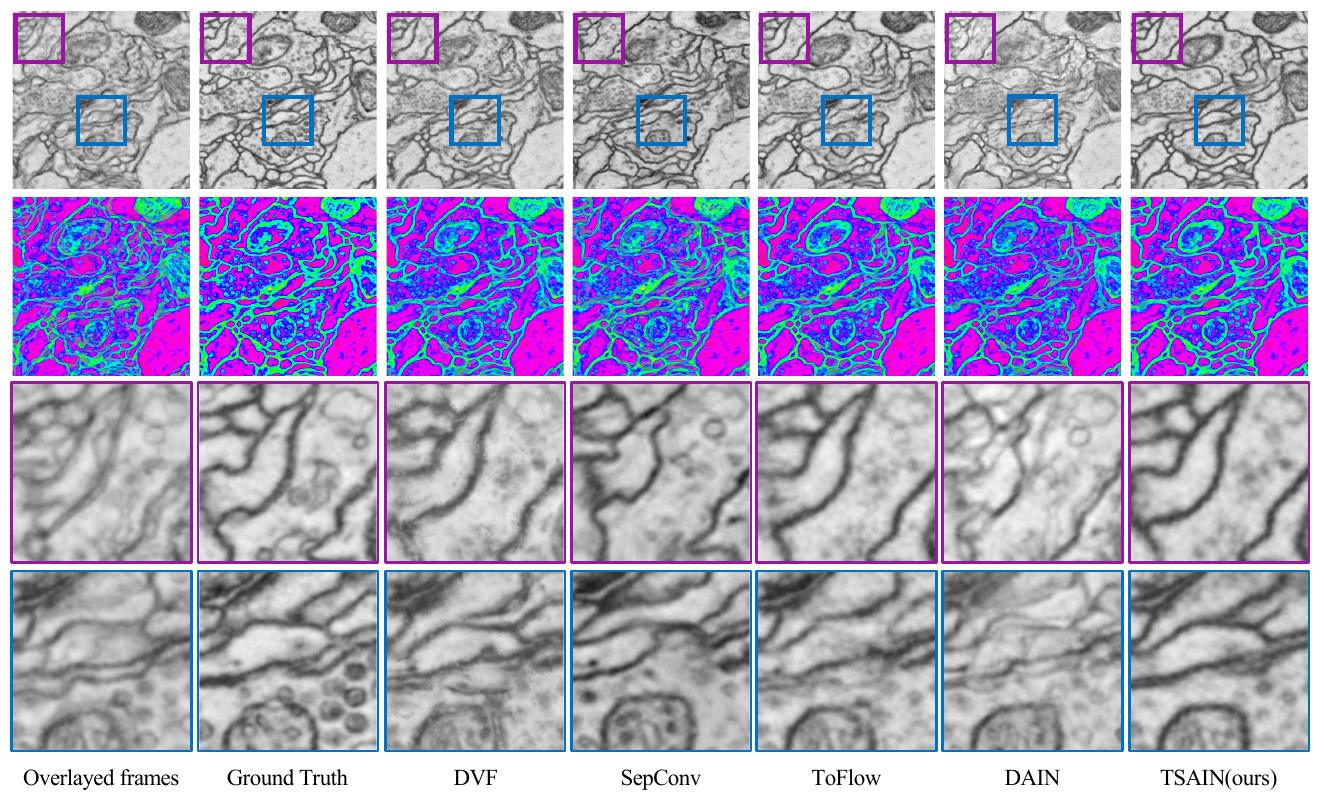}
   \caption{Visual comparisons on the cremi$\_$triplet B. The second line is the result of fusing two images in the HSV color space. The more dark regions, the worse the performance.}
\label{fig:cmp}
\end{figure}

\subsection{Comparison to State-of-the-art Methods}
We compare the performance of our proposed method to several state-of-the-art (SOTA) interpolation approaches on the processed CREMI datasets. For fair comparison, we adjust the loss function of both the proposed method and the compared approaches to L$_{total}$ and conduct quantitative experiments under the same experimental environment. Qutitative results are shown in Table~\ref{tab:cmps}. We learn the following facts: (1) ToFlow is the best performing VFI approach among the compared four methods although the amount of model parameters is minimal; (2) Complex flow modeling does not work well on EM images. Despite a large number of model parameters, DAIN performs even worse than SepConv. As the model becomes more complex, the errors caused by incorrect optical flow can be accumulated. In addition, we can see that our framework outperforms ToFlow by 0.37dB on cremi$\_$triplet A, 0.24dB on cremi$\_$triplet B, and 0.59dB on cremi$\_$triplet C in terms of PSNR. Compared with ToFlow, our method has a slightly larger model size but is nearly 2 times faster.

Visual results of different methods are illustrated in Figure~\ref{fig:cmp}. We see that the proposed network can generate visually pleasing EM frames with more fine details, more continuous edges, more accurate structures and fewer blurry artfacts even for challenging EM scenes. Significant improvements and visual-pleasing results on EM images demonstrate our network can handle more complicated Spatio-temporal patterns and unstable image quality than other compared methods.

\subsection{Ablation Study}
To further illustrate the effectiveness of different components in our model, we make a comprehensive ablation study.

\begin{table}[ht]
\centering
\resizebox{0.47\textwidth}{!}{
\begin{tabular}[b]{cccccc}
\toprule
\multirow{2}{*}{Methods}
 & w/ residual spatial- & w/ deformable & \multirow{2}{*}{PSNR} & \multirow{2}{*}{SSIM} & \multirow{2}{*}{IE} \\
  & adaptive blocks &  refinement blocks &  &  & \\

\midrule
\midrule
$Baseline$ & - & - & 17.15 & 0.3874 & 26.32 \\ 
\midrule
 $M_{1}$ & $\checkmark$ & - & 17.27 & 0.3946 & 26.02 \\
 $M_{2}$ & - & $\checkmark$ & 17.34 & 0.3978 & 25.74 \\
 $M_{3}$ & $\checkmark$ & $\checkmark$ & 17.52 & 0.4042 & 25.29 \\

\bottomrule
\end{tabular}}
\caption{Ablation study on the proposed modules. We calculate the PSNR, SSIM and IE on cremi$\_$triplet B.}
\label{tab:ab1}
\end{table}

\begin{table}[ht]
\centering
\resizebox{0.47\textwidth}{!}{
\begin{tabular}[b]{cccccccc}
\toprule
\multirow{2}{*}{Methods} & \multicolumn{4}{c}{Number of DRBs} & \multirow{2}{*}{PSNR} & \multirow{2}{*}{SSIM} & \multirow{2}{*}{IE} \\
& Zero & One & Two & Three &  &  & \\
\midrule
\midrule
$Baseline$ & $\checkmark$ &       -      &       -      &       -      & 17.27 & 0.3946 & 26.32 \\ 
\midrule
 $N_{1}$ &       -      & $\checkmark$ &       -      &       -      & 17.22 & 0.3837 & 26.15 \\
 $N_{2}$ &  	 -    	&  		-	   & $\checkmark$ &       -      & 17.41 & 0.3939 & 25.58 \\
 $N_{3}$ &       -      &       -      &       -      & $\checkmark$ & 17.52 & 0.4042 & 25.29 \\
\bottomrule
\end{tabular}}
\caption{Ablation study on the cremi$\_$triplet B about the number of stackable deformable refinement blocks.}
\label{tab:ab2}
\end{table}

\begin{figure}[ht]
\centering
\includegraphics[width=0.48\textwidth]{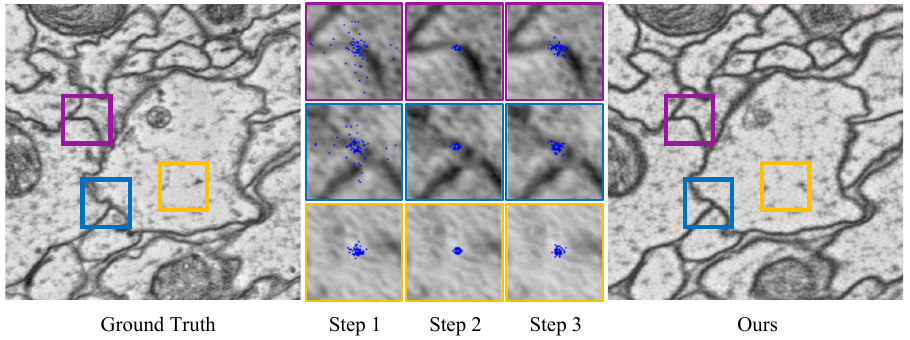}
   \caption{Visualization of the learned sampling points. The steps from 1 to 3 demonstrate the effectiveness of RSABs in a coarse-to-fine manner.}
\label{fig:rsab}
\end{figure}

\begin{figure}[ht]
\centering
\includegraphics[width=0.45\textwidth]{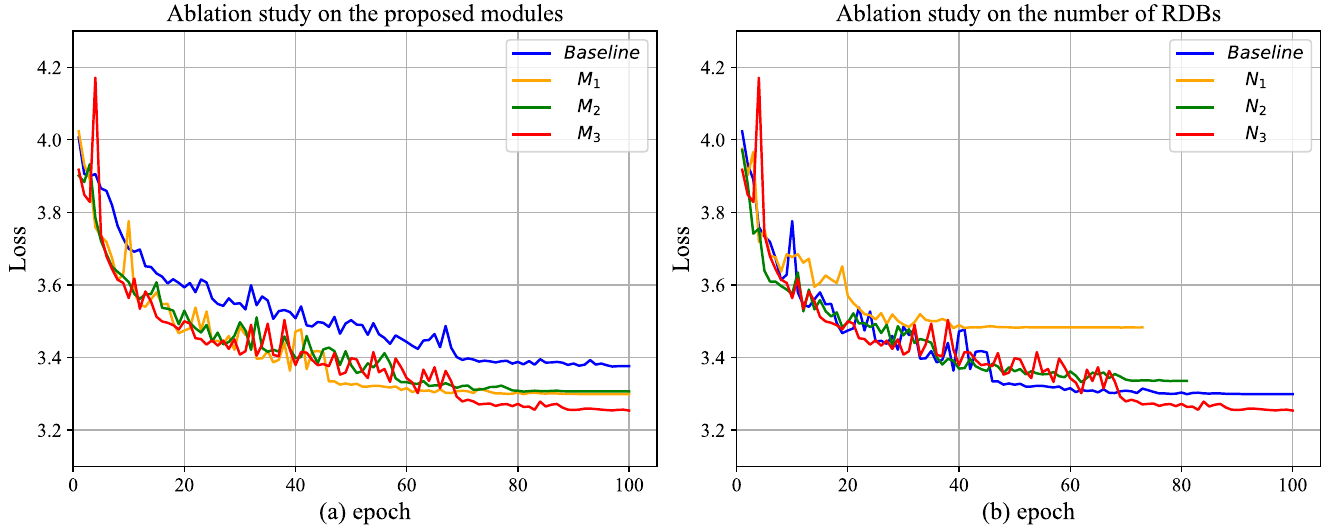}
   \caption{Loss to ablation studies on the proposed modules and the number of DRBs, respectively.}
\label{fig:abloss}
\end{figure}

\paragraph{Effectiveness of Residual Spatial-Adaptive Module.}
As shown in Table~\ref{tab:ab1}, the residual spatial-adaptive module achieves a significant PSNR improvement from 17.15dB to 17.27dB. Figure~\ref{fig:abloss} (a) illustrates the effectiveness of the RSABs from the comparison of $M_{1}$ and $baseline$. Furthermore, we visualize the learned offsets of the RSABs in Figure~\ref{fig:rsab}. From Step 1, We can see that in the smooth areas or uniform texture areas, the sampling points are closely distributed in clusters, while at the locations close to the edge, the sampling points extend along the edge. Most of them fall on positions that have similar textures to the green dots. The sampling tends to be stable from Step 1 to Step 3, and the sampling points are tightly clustered. It shows that our RSABs can enhance the edge's continuity and handle unstable image quality with spatial adaptive sampling.

\paragraph{Effectiveness of Deformable Refinement Blocks.}
The DRBs aim to further refine and generate more precise intermediate frames through a feedback mechanism. We can observe from Table~\ref{tab:ab1} that the module with deformable refinement achieves 0.19dB improvement compared with the baseline. Moreover, we explore the influence of the number of stacked DRB modules on the interpolation in Table~\ref{tab:ab2} and Figure~\ref{fig:abloss}. As the number of the stacked DRBs increases, the results of frame interpolation performs better.

\section{Conclusion}
In this paper, we propose an effective EM image interpolation method to handle complex deformation and unstable image quality. Our proposed temporal spatial-adaptive module adaptively aggregates spatial-related pixels and enhances edge continuity using neighboring similar textures in temporal features. Furthermore, we introduce a stackable deformable refinement module to extract aligned relevant features from the input reference frames, which performs the feedback mechanism under the supervision of the input frames. Extensive experiments demonstrate our proposed approach achieves state-of-the-art performance on CREMI datasets compared with flow-based methods.

\section*{Acknowledgments}
This work was supported in part by A (a), B (b), and C (c).

\clearpage

\bibliographystyle{named}
\bibliography{ijcai21}

\end{document}